%% file: main_icra.tex
\documentclass[letterpaper, 10 pt, conference]{ieeeconf}  %

\IEEEoverridecommandlockouts
\usepackage{cite}
\usepackage{amsmath,amssymb,amsfonts}
\usepackage{algorithmic}
\usepackage{graphicx}
\usepackage{textcomp}
\usepackage{xcolor}
\def\BibTeX{{\rm B\kern-.05em{\sc i\kern-.025em b}\kern-.08em
    T\kern-.1667em\lower.7ex\hbox{E}\kern-.125emX}}

\usepackage{url}
\usepackage{hyperref} 
\hypersetup{hidelinks,
    colorlinks=true,
    allcolors=black,
    pdfstartview=Fit,
    breaklinks=true
}
\usepackage{siunitx}
\usepackage{threeparttable}
\usepackage{multirow}
\usepackage{makecell}
\usepackage{booktabs}
\usepackage{float}
\usepackage{gensymb}
\usepackage{hyperref}
\hypersetup{colorlinks=false}

\usepackage{tikz}
\newcommand*\circled[1]{\tikz[baseline=(char.base)]{
            \node[shape=circle,draw,inner sep=1pt,font=\small] (char) {#1};}}

\begin{document}
\title{
RAG-RUSS: A Retrieval-Augmented \\Robotic Ultrasound for Autonomous Carotid Examination
}

\author{Dianye Huang$^{1,2,*}$, Ziping Cong$^{1,*}$, Nassir Navab$^{1,2}$, \textit{Fellow, IEEE}, and Zhongliang Jiang$^{1,2,^\dagger}$
\thanks{$^*$ Contribute Equally. $^\dagger$ Corresponding author(e-mail: zl.jiang@tum.de).}
\thanks{$^1$ Computer-Aided Medical Procedures and Augmented Reality (CAMP), Technical University of Munich (TUM), Germany,}
\thanks{$^2$  Munich Center for Machine Learning (MCML), Germany}
}

\maketitle

\begin{abstract}
Robotic ultrasound (US) has recently attracted increasing attention as a means to overcome the limitations of conventional US examinations, such as the strong operator dependence. However, the decision-making process of existing methods is often either rule-based or relies on end-to-end learning models that operate as black boxes. This has been seen as a main limit for clinical acceptance and raises safety concerns for widespread adoption in routine practice. To tackle this challenge, we introduce the RAG-RUSS, an interpretable framework capable of performing a full carotid examination in accordance with the clinical workflow while explicitly explaining both the current stage and the next planned action. Furthermore, given the scarcity of medical data, we incorporate retrieval-augmented generation to enhance generalization and reduce dependence on large-scale training datasets. The method was trained on data acquired from 28 volunteers, while an additional four volumetric scans recorded from previously unseen volunteers were reserved for testing. The results demonstrate that the method can explain the current scanning stage and autonomously plan probe motions to complete the carotid examination, encompassing both transverse and longitudinal planes.

\end{abstract}

\input{intro}

\input{text}

\bibliographystyle{IEEEtran}
\bibliography{IEEEabrv, reference}

\end{document}

%% file: intro.tex
\section{Introduction}
\label{sec:intro}
\par
Ultrasound (US) is a non-invasive, real-time imaging modality with wide accessibility, establishing it as the primary diagnostic tool in contemporary clinical practice~\cite{izadifar2017mechanical}. However, the quality of US imaging and the efficiency of acquiring standard imaging planes are highly dependent on the operator’s proficiency in probe manipulation. This reliance on operator expertise introduces variations in examination outcomes and requires years of training to attain proficiency, which exacerbates workforce shortages, especially in under-resourced regions~\cite{sippel2011use}. 

\par
In order to reduce physician workloads and examination variability, over the past two decades, the autonomous robotic ultrasound scanning system (RUSS) has been extensively researched and is receiving ongoing attention~\cite{jiang2023robotic}. Various RUSS have been developed for a broad range of clinical applications, including autonomous screening of the breast~\cite{welleweerd2021out}, lung~\cite{ma2021autonomous}, and blood vessels~\cite{huang2023motion, huang2024robotassist}. These systems mainly focus on automating the screening process by leveraging US imaging feedback and/or external RGB-D cameras. Wang~\emph{et al.}~\cite{wang2024autonomous} proposed a vision-servo-based autonomous carotid US scanning system using an improved Siamese network to track the target vessel. Huang~\emph{et al.}~\cite{huang2024robot} proposed a two-stage process for automatic switching between transverse and longitudinal sections, combining impedance/force control with image segmentation. However, these systems mainly focus on motion control and rely heavily on rule-based policies, which struggle to exhibit higher-level understanding and dynamic decision-making capabilities.

\begin{figure}[!t]
  \centering
  \includegraphics[width=0.8\columnwidth]{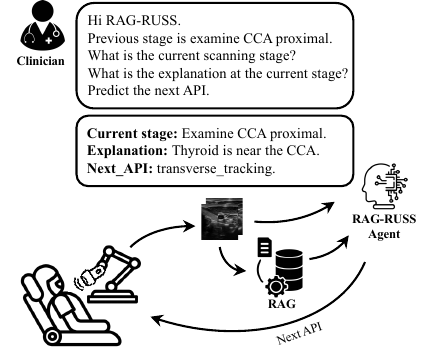}
  \caption{An illustration showing a representative use case of the RAG-RUSS system.
  Given a clinician’s query, RAG-RUSS retrieves similar annotated scans via RAG, then identifies the \texttt{stage}, generates an \texttt{explanation}, and predicts the \texttt{next\_API}. This API is performed by the robotic arm, and the acquired US images are fed back into the system, forming a closed loop that enables autonomous, interpretable carotid artery scanning.}
  \vspace{-0.5cm}
  \label{fig:teaser}
\end{figure}

\par
To address these limitations, learning-based RUSS have emerged as a promising direction, enabling more intelligent, data-driven perception and control~\cite{bi2024machine}. In order to reduce the reliance on handcrafted rules, Huang~\emph{et al.}~\cite{huang2021towards} introduce clinical imitation learning (IL) frameworks to replicate the operational habits of experienced physicians. Given the current observation, reinforcement learning (RL) is also employed to infer the next control commands to maneuver the US probe toward the standard imaging plane~\cite{jiang2024intelligent, su2024fully, li2021autonomous}. To address inter-operator variability in US imaging, Jiang~\emph{et al.} propose an intelligent robotic sonographer that autonomously navigates to standard imaging planes by learning expert-level scanning strategies. In their work, a neural reward function, trained via ranked pairwise image comparisons in a self-supervised manner, captures high-level physiological knowledge. Despite the promising results achieved by the aforementioned approaches, the practical deployment of RUSS remains hindered by the inherent opacity of IL and RL paradigms. These data-driven methods typically encode expertise through only image–trajectory demonstration pairs, yielding black-box policies that mirror the opacity of human decision-making. Transparent system designs that allow physicians to monitor the internal states of trained policies have therefore become essential~\cite{song2025intelligent}. Building upon this need for policy transparency, language emerges as a natural bridge connecting the medical robots, physicians, and patients in real-world clinical settings. It is worth noting that achieving such embodied communicative capability necessitates the integration of low-level perceptual inputs with high-level semantic reasoning, an area in which traditional rule-based and IL/RL approaches remain limited~\cite{kim2025srt, roy2021machine}. 

\par
Recently, large language models (LLMs) have shown strong associative capabilities and broad commonsense knowledge, accelerating progress in embodied artificial intelligence~\cite{liu2025aligning, driess2023palm}. Motivated by this progress, the research community starts exploring the use of vision-language models (VLMs) and vision-language-action (VLA) frameworks to enable intelligent, interactive agents. Wang~\emph{et al.} propose EndoChat, a specialized multimodal large language model (MLLM) designed for diverse dialogue tasks in robotic-assisted surgery, trained on the Surg-396K dataset. Ng~\emph{et al.}~\cite{ng2025endovla} introduce EndoVLA, a VLA model tailored for continuum robots in gastrointestinal interventions, capable of semantic polyp tracking, delineating abnormal mucosal regions, and following circular cutting markers. 
In the field of US, Xu \emph{et al.} introduced USPilot~\cite{chen2025uspilot}, an embodied, LLM-powered robotic US assistant that acts as a virtual sonographer—answering patient queries, interpreting US-specific tasks through fine-tuning, and invoking APIs based on user intent. The system integrates a fine-tuned LLM with an LLM-enhanced graph neural network for API control and task planning. However, its primary focus lies in intent understanding and high-level planning, with limited capability for real-time image content analysis, constraining its effectiveness in dynamic task execution. Alternatively, Jiang~\emph{et al.}~\cite{jiang2025towards} proposed UltraBot, a carotid artery ultrasonography system trained on a large-scale dataset comprising $247$ k image–action expert demonstrations. UltraBot employs end-to-end imitation learning to directly map real-time US images to 6-DoF probe motions, while integrating scanning, biometric measurement, plaque segmentation, and report generation into a unified workflow. However, UltraBot's decision-making process remains a black box without explicit modeling of intermediate reasoning steps or stage-wise task decomposition.
While large model-based methods have shown promising results, they require extensive domain-specific training datasets, which are often prohibitively expensive to collect. To mitigate this limitation, and inspired by RAG-Driver~\cite{yuan2024rag} from the autonomous driving domain, we propose incorporating a retrieval-augmented generation (RAG) component to reduce dependency on large training datasets and alleviate data scarcity bottlenecks in the medical field. 

\par
In this work, we introduce RAG-RUSS, a retrieval-augmented framework for carotid US scanning that couples a VLM with an RAG component to reduce dependence on large training corpora. Unlike systems such as USPilot and UltraBot, which emphasize intent understanding or direct end-to-end action mapping, our approach leverages retrieved, similar scan contexts to enable image understanding and explanation-driven decision-making during closed-loop execution. To train RAG-RUSS, we also constructed a high-quality dataset from 32 human volumetric scans. Within a clinically defined workflow, RAG-RUSS predicts the current stage and provides stepwise explanations alongside the next\_API. This design supports standardized carotid scanning by adapting evidence from prior cases while maintaining transparency throughout the examination. To the best of our knowledge, this is the first robotic ultrasound system to integrate perception, explanation, and action in a unified framework for carotid artery examination. By explaining both the current image and the subsequent action, RAG-RUSS takes an important step toward trustworthy, clinically acceptable deployment.

\par
The rest of this paper is organized as follows: \textit{Section~\ref{sec:II_data}} outlines the clinical workflow for carotid artery scan and the data preparation process. \textit{Section~\ref{sec:method}} presents the architecture of the proposed RAG-RUSS. \textit{Section~\ref{sec:expres}} describes the experimental setup and reports the results. Finally, \textit{Section~\ref{sec:conclusion}} concludes the study.

\begin{figure*}[!t]
  \centering
  \includegraphics[width=0.7\textwidth]{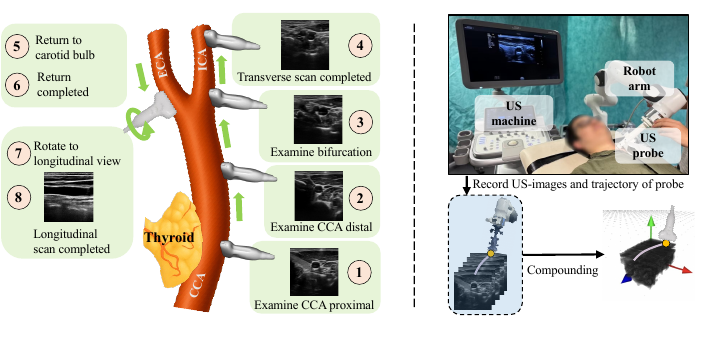}
  \vspace{-0.3cm}
  \caption{Overview of the carotid ultrasound scanning workflow and data acquisition/processing pipeline. \textit{Left:} presents eight predefined scanning stages of the carotid artery examination. \textit{Right:} US images and the probe’s pose trajectory are recorded and subsequently compounded into a 3D volumetric data for use in the simulation environment. 
  }
  \vspace{-0.3cm}
  \label{fig:workflow}
\end{figure*}

%% file: text.tex
\section{Examination Workflow and Data Preparation}
\label{sec:II_data}
\subsection{Carotid Ultrasound Examination Workflow}
\subsubsection{Clinical Workflow for Carotid Artery Examination}
\par
The carotid arteries, located on either side of the trachea at the neck, are the primary vessels that deliver oxygen-rich blood from the heart to the brain and face. Each carotid artery comprises a common carotid artery (CCA) that bifurcates into the internal carotid artery (ICA), supplying blood to the brain and eyes, and an external carotid artery (ECA), supplying blood to the face and scalp (see Fig.~\ref{fig:workflow}). Atherosclerosis is one of the most prevalent carotid diseases, in which plaque accumulation leads to carotid stenosis (luminal narrowing) and elevates stroke risk, particularly for the aged population. In this context, medical US has become the standard first-line diagnostic tool due to its advantages of being radiation-free and providing real-time imaging.

\par
For regular carotid examination, sonographers generally follow a standard workflow~\cite{huang2021towards,huang2024robot}. To ensure full satisfaction of the clinical diagnostic purpose, sonographers need to fully cover the CCA and its bifurcation into the internal and external branches. The examination is typically performed first in the transverse (short-axis) plane along the lumen centerline and then in the longitudinal (long-axis) plane of the carotid artery~\cite{wang2024autonomous}. The scan usually begins in transverse view at the proximal CCA, proceeds distally to the carotid bulb (carotid sinus), and continues across the bifurcation to visualize the ICA and ECA. The physician then switches to longitudinal views for detailed assessment and measurements, such as the vessel diameter.

\subsubsection{Scanning Stage Definition}
\par
To automate the scanning procedure and facilitate the holistic understanding of the overall progress of the examination for RUSS, we adopt the above clinical workflow to an automatic system, following~\cite{huang2024robot, jiang2025towards}. In this study, we partition the carotid scan process into eight sequential stages (see Fig.~\ref{fig:workflow}), each guided by distinct anatomical cues from the US images. 
\begin{itemize}
\item[\circled{1}] \texttt{Examine CCA proximal:} the CCA is visualized together with the thyroid gland.
\item[\circled{2}] \texttt{Examine CCA distal:} the CCA remains visible, but the thyroid gland is absent from the images.
\item[\circled{3}] \texttt{Examine bifurcation:} the carotid bulb is reached; the lumen starts diverging into ECA and ICA.
\item[\circled{4}] \texttt{Transverse scan completed:} the lumen is clearly divided into the ICA and ECA.
\item[\circled{5}] \texttt{Return to carotid bulb:} as the probe is moved back toward the bifurcation, the ICA and ECA gradually converge into a single lumen.
\item[\circled{6}] \texttt{Return completed:} the ICA and ECA converge into the common carotid artery at the carotid bulb.
\item[\circled{7}] \texttt{Rotate to longitudinal view:} the artery cross-section changes from an elliptical shape to an elongated, rectangular profile.
\item[\circled{8}] \texttt{Longitudinal scan completed:} the longitudinal cross-section of carotid artery is clearly visualized. 
\end{itemize}

\subsection{Ultrasound Simulation Using Human Volumetric Data}
\par
Due to safety concerns, it is impractical to collect all the required training scanning trajectories directly from volunteers. Following previous studies~\cite{li2021autonomous}, we built a simulation environment based on recorded human volumetric data, in which a virtual probe is initialized to perform continuous scanning. Each paired set of simulated US images and the corresponding probe trajectory is regarded as one demonstration of a complete carotid scan. Notably, a single pre-recorded volume can generate multiple demonstrations. To mitigate the risk of overfitting, a total of $32$ three-dimensional US volumes of the carotid artery were acquired and reconstructed from scans of $32$ healthy volunteers.



\subsubsection{Human Volumetric Data Acquisition}
\par
The data acquisition platform shown in Fig.~\ref{fig:workflow} was built on a Panda robotic arm (Franka Emika, Germany) integrated with an ACUSON Juniper ultrasound (US) system (Siemens, Germany). A linear US probe (12L3, Siemens, Germany) was mounted on the robot end-effector using a custom-designed, 3D-printed probe holder. The US images were captured via a frame grabber (USB Capture HDMI, MAGEWELL, China) at a frequency of 30~Hz. Meanwhile, the robotic tracking data were recorded through a ROS interface at a higher rate of approximately 100~Hz to compensate for potential temporal misalignment between the two data streams. By stacking the 2D US images with their corresponding robot poses in 3D space, a reconstructed 3D US volume can be generated using a classic interpolation technique such as PLUS~\cite{lasso2014plus}. An illustrative example of the reconstructed volume is shown in Fig.~\ref{fig:workflow}. A total of 32 carotid volumes were acquired from 32 healthy volunteers. Each volume was manually checked to ensure that the vascular morphology was clearly visible and that the bifurcation structures were fully captured. 
To show the variety in the group, the volunteers’ demographic and physical information is summarized in Table~\ref{tab:demographics}.

\par
The compounded carotid volume is visualized using OpenGL. By initializing a virtual US probe, we can mimic the standard examination workflow discussed in the previous section to generate the carotid examination demonstration with paired probe pose and B-mode image. By adjusting the virtual probe position, a set of standardized carotid examination demonstrations can be obtained. Such a human volumetric data-based simulation preserves the fidelity of medical ultrasound images while providing a flexible and practical controllable scanning trajectory generation. 

\begin{table}[!h]
\centering
\caption{statistics of volunteers (N=32).}
\label{tab:demographics}
\resizebox{0.43\textwidth}{!}{%
\begin{tabular}{ccccc}
\toprule
\textbf{Age (years)} & \textbf{Height (m) }& \textbf{Weight (kg) }& \textbf{BMI} \\
\midrule
27.97 $\pm$ 3.91 & 1.74 $\pm$ 0.09 & 70.22 $\pm$ 12.24 & 23.06 $\pm$ 2.86 \\
\bottomrule
\end{tabular}
}
\end{table}

\subsubsection{Expert Carotid Examination Demonstration Generation}
\par
This section introduces the implementation details for generating expert demonstrations of carotid examination using the reconstructed human volumetric data. First, a set of anatomical waypoints corresponding to different scanning stages was manually defined based on visual vessel features. To standardize the scanning trajectory generation, we further compute the centroid point of the vessel in the labeled binary vessel mask to have a refined scanning trajectory. Then, a scanning trajectory can be generated by connecting these refined waypoints and adjusting its orientation in specific stage transitions. By automatically executing this continuous scanning trajectory in the simulation, the paired B-mode image can be obtained for individual probe locations.  

\par
For each reconstructed volume, three complete scans were conducted from the proximal segment of the CCA to the longitudinal plane visualization, as illustrated in Fig.~\ref{fig:workflow}. To balance the sample distribution across different scanning stages, an additional 20 scans were acquired for individual stages with relatively fewer samples, including \textit{Transverse scan completed}, \textit{Return completed}, and \textit{Longitudinal scan completed}. During the transverse scanning stage, the probe was advanced in increments of 1 mm per step. In the return stage, the first step retracted 2 mm, followed by subsequent steps of 1.5 mm each. To introduce variation among scans obtained from the same volunteer, random perturbations were added to the motion commands at each step. Specifically, translational deviations of up to $\pm0.3$ mm in the longitudinal direction and $\pm0.2$ mm in the lateral direction were applied.
\subsection{Dataset Structure}
\par
To train the proposed RAG-RUSS with the ability to interact with human users, the dataset was designed to incorporate both visual and textual information. A sliding time window of N steps was employed. For each entry within the window, in addition to the US images, we defined three types of textual annotations: (1) the current scanning stage, such as Stage 1 \textit{Examine CCA proximal} and Stage 2 \textit{Examine CCA distal} (see Fig.~\ref{fig:workflow}); (2) an explanation of the key anatomical feature characterizing the stage; and (3) the name of the next API required to complete the given task. The stage-specific explanations were provided by our clinical collaborator according to standard criteria used to differentiate scanning stages during carotid examinations. For example, the explanations for Stages 1 and 2 are ``Thyroid is near the CCA” and ``Thyroid is not visible,” respectively. In this study, three APIs were defined: ``tracking forward,” ``tracking backward,” and ``rotation clockwise,” which correspond to advancing the probe, retracting the probe, and rotating the probe from the transverse plane to the longitudinal plane, respectively. In total, we collected 15,459 items (N-length sliding window), inlcuding paired visual and textual information based on 32 reconstructed volumes. The images are in the resolution of $224 \times 224$ pixels. To avoid data leakage, four out of 32 volunteers' data is kept for testing. 

For different training objectives, three sub-datasets were constructed: (1) dataset A for training the RAG module, (2) dataset B for pre-training the cross-modality projector (2-layer MLP) to align visual and textual features, and (3) dataset C for fine-tuning the cross-modality MLP together with the VLM backbone using LoRA adapters~\cite{hu2022lora}. Dataset A consisted of the first and last images within each moving window, along with the textual stage description corresponding to the last frame. In total, 12,937 entries from 28 volunteers were included. For training, we iterated over each entry and generated 20 paired positive and negative samples to support contrastive learning. Dataset B was constructed from individual B-mode images paired with predefined stage-specific queries and corresponding answers that describe the anatomical features characterizing each stage. Dataset C was designed to refine the LLM backbone and includes multi-turn question–answer pairs concerning the current stage, the next API, and related instructions, thereby enabling effective interaction with human users.

\section{Method}
\label{sec:method}
\par
In this section, we elaborate on the structure and the training pipeline of RAG-RUSS, an intelligent sonography agent that explains its current scanning stage at each timestep, enabling the physician to monitor progress and intervene if the procedure deviates from the plan. Effective, natural communication between the agent and the physician is therefore essential. To this end, RAG-RUSS is built around an LLM to leverage its learned ``common sense’’ and intrinsic capacity for natural-language communication, while seamlessly incorporating physician expertise.

\begin{figure*}[!t]
  \centering
  \includegraphics[width=0.95\textwidth]{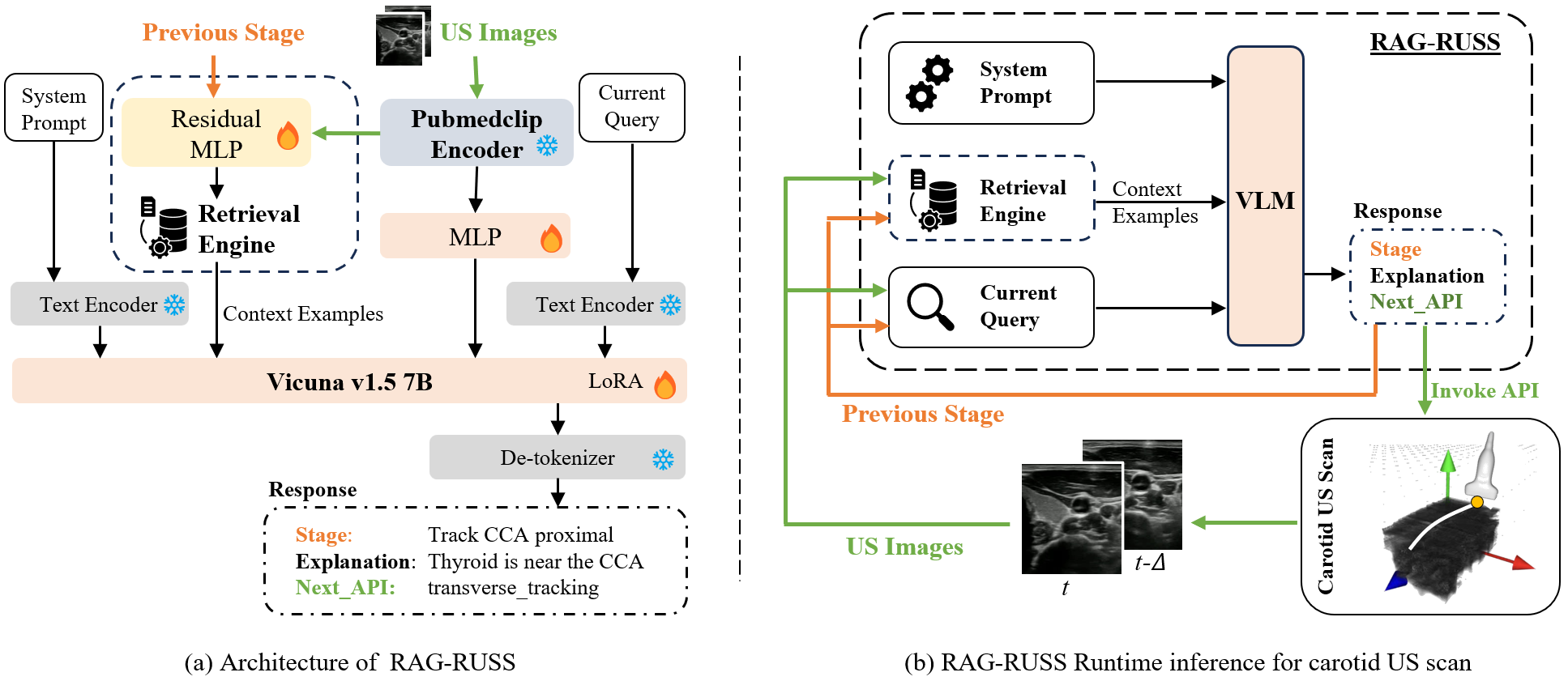}
  \vspace{-0.3cm}
  \caption{Architecture and runtime inference of the proposed RAG-RUSS for carotid artery US scanning. \textit{Left:} System architecture comprising an LLM backbone (Vicuna v1.5, 7B~\cite{zheng2023judging}), the medical vision foundation model (PubMedCLIP-ViT-B/32~\cite{eslami2021does}), and a RAG module that retrieves similar scanning contexts to support decision-making. \textit{Right:} Inputs/outputs of RAG-RUSS and the signal flow when deploying it for carotid US scanning.  For more details on inputs/outputs and architecture of RAG-RUSS, refer to \textit{Fig.~\ref{fig:prompts}} and \textit{Section~\ref{sec:method}-B}, respectively.
  }
  \vspace{-0.3cm}
  \label{fig:overview}
\end{figure*}

\subsection{Structure of RAG-RUSS}
\par
\begin{figure}[!t]
  \centering
  \includegraphics[width=0.485\textwidth]{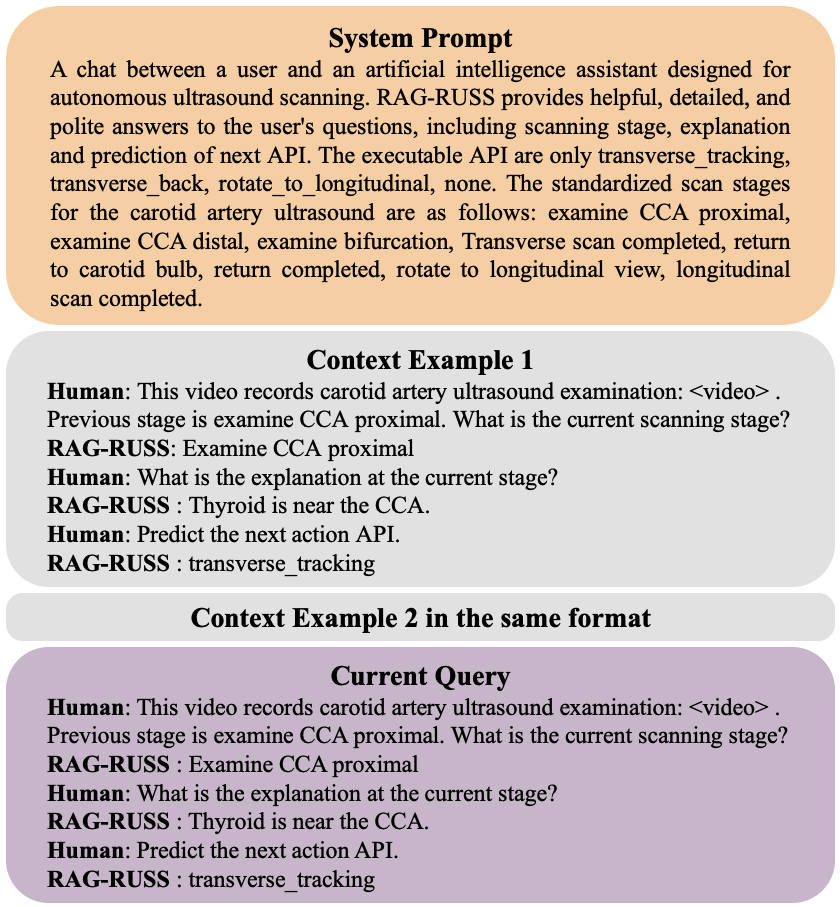}
  \vspace{-0.6cm}
  \caption{An illustrutive multi-turn QA example of RAG-RUSS. The inputs are: i). System prompt that specifies the task description along with the predefined executable APIs and scanning stages. ii). Two retrieved scanning contexts via the RAG component. iii). The current query involves the two input US images and the previously predicted stage, and then poses three questions sequentially. RAG-RUSS then outputs: i). the current \texttt{stage}, ii). a short \texttt{explanation}, and iii). the \texttt{next\_API} to execute.
  }
  \vspace{-0.4cm}
  \label{fig:prompts}
\end{figure}

As illustrated in Fig.~\ref{fig:overview}~(a), RAG-RUSS comprises an LLM, a vision foundation model trained on large-scale medical data, and a retrieval engine that extracts relevant knowledge to support decision-making. In Fig.~\ref{fig:overview}~(b), during inference, RAG-RUSS follows a multi-turn question–answering workflow to address three queries: identifying the current \texttt{stage}, generating an \texttt{explanation}, and predicting the \texttt{next\_API}. We also incorporate multimodal in-context learning (ICL), enabling the model to reference visual–linguistic history and thereby improve scanning accuracy and reliability. At each query step, RAG-RUSS produces interpretable text, ensuring transparency throughout the scan (see an illustrative multi-turn QA example in Fig.~\ref{fig:prompts}.) 


\subsubsection{Large Language Model}
\par
Vicuna v1.5 7B~\cite{zheng2023judging} is employed as the backbone LLM. Here, Vicuna takes as input the visual embeddings from a vision encoder, the system and query prompts encoded by the text encoder, and the retrieval-augmented historical context. It produces three types of outputs: the current scanning stage \texttt{stage} (see the stage definitions in \textit{Section~\ref{sec:method}-A}), an explanation of that stage \texttt{explanation}, and the prediction of the \texttt{next\_API}. 


\subsubsection{Ultrasound Image Encoder}
\par
A frozen PubMedCLIP-ViT-B/32~\cite{eslami2021does} is adopted as the vision encoder for RAG-RUSS. This encoder is pretrained on a large-scale biomedical multimodal dataset ROCO~\cite{ruckert2024rocov2}, which enables it to effectively capture texture, boundary, and structural features in medical imaging modalities such as US, X-ray, and MRI. The input consists of two US frames [the most recent US image $\mathbf{I}_{t}\in\mathbb{R}^{224\times224}$ captured at timestep $t$, and the previous US image $\mathbf{I}_{t-\Delta}$ captured at timestep $t-\Delta$, as illustrated in Fig.~\ref{fig:overview}~(a)]. Each image is encoded by PubMedCLIP into a sequence of patch-level tokens, where the \texttt{[CLS]} token \textit{from the second-last layer} is fed into the RAG module, while the remaining visual tokens \textit{from the same layer} are passed to the LLM. 

\subsubsection{Cross-Modality Projector}
\par
Following the LLaVA~\cite{zheng2023judging} design, we employed a two-layer MLP as a cross-modality projector to align visual and textual features. The non-linear activation function $\sigma(\cdot)$ (GELU in this work) is applied after each linear layer.  Formally, given the visual embedding $z \in \mathbb{R}^{98 \times 768}$, the projection is defined as:
\begin{equation}
z_{v} = \sigma \big( W_2 \cdot \sigma ( W_1 \cdot z ) \big),
\label{eq:proj}
\end{equation}
where $W_1$ and $W_2$ are learnable weight matrices. The output $z_{v} \in \mathbb{R}^{98 \times 4096}$ is thus aligned with the LLM's token-embedding space. This projector is trained jointly during both the pretraining and fine-tuning. 

\subsubsection{RAG-based In-Context Learning}
\par
\begin{figure}[!tpb]
  \centering
  \includegraphics[width=0.48\textwidth]{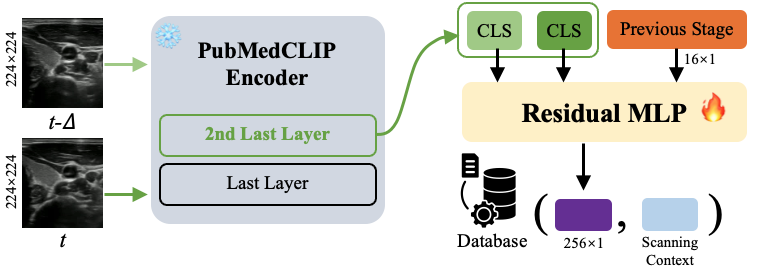}
  \vspace{-0.2cm}
  \caption{Scheme for building an RAG database. The scanning context includes two US frames, the previous stage, and associated VQA annotations.}
  \label{fig:rag_resmlp}
\end{figure}

To perform the retrieval of historical scanning context, including two US frames, the previous stage, and VQA annotations (see the Context Example in Fig.~\ref{fig:prompts}), we first train a residual MLP (ResMLP) to project the context information into the same embedding space. As illustrated in Fig.~\ref{fig:rag_resmlp}, the input consists of two \texttt{[CLS]} tokens extracted from two US images (each a 768×1 vector) and the textual embedding of the previous stage. The textual previous stage is mapped into a 16-dimensional embedding using Table~\ref{tab:stage_embeddings}. These three embeddings are concatenated into a 1552-dimensional vector, which is then passed through the ResMLP to project it into a 256-dimensional embedding space. A triplet loss in Eq. (\ref{eq:triplet_loss}) is utilized to train the ResMLP.
\begin{equation}
\label{eq:triplet_loss}
\mathcal{L}_{\text{tri}}(\mathbf{a},\mathbf{p},\mathbf{n}) = \max \left( \lVert \mathbf{a} - \mathbf{p} \rVert_{2} - \lVert \mathbf{a} - \mathbf{n} \rVert_{2} + \beta, \, 0 \right)
\end{equation}
where $\mathbf{a}$, $\mathbf{p}$, $\mathbf{n}$, and $\beta=0.75$ denote the anchor, positive, negative samples, and marginal distance, respectively. 

\par
During the retrieval, the resulting embedding is compared against a database of historical scanning contexts to compute cosine similarity scores. The top-$k$ most similar scanning contexts ($k=2$ in this work) are retrieved and supplied to the LLM as additional context. This retrieval mechanism allows the model to reference clinically similar scenarios at query time, improving stage-recognition accuracy and overall scanning success.

\begin{table}[!t]
    \centering
    \caption{Textual Stage to embedding vector mapping.}
\resizebox{0.4\textwidth}{!}{%
    \begin{tabular}{lc}
    \toprule
    \multicolumn{1}{c}{\textbf{Stage}} & \multicolumn{1}{c}{\textbf{16-dim. Embedding Vector}} \\
    \midrule
    Examine CCA proximal & [0.1, 0.1, 0.1, $\cdots$, 0.1] \\
    Examine CCA distal   & [0.2, 0.2, 0.2, $\cdots$, 0.2] \\
    Examine bifurcation  & [0.4, 0.4, 0.4, $\cdots$, 0.4] \\
    Transverse completed     & [0.5, 0.5, 0.5, $\cdots$, 0.5] \\
    Return to carotid bulb     & [0.6, 0.6, 0.6, $\cdots$, 0.6] \\
    Return completed    & [0.7, 0.7, 0.7, $\cdots$, 0.7] \\
    Rotate to longitudinal view & [0.8, 0.8, 0.8, $\cdots$, 0.8] \\
    Longitudinal scan completed      & [0.9, 0.9, 0.9, $\cdots$, 0.9] \\
    \bottomrule
    \end{tabular}
}
    \vspace{-0.2cm}
    \label{tab:stage_embeddings}
\end{table}

\subsection{Training Pipeline}
\par
The training of RAG-RUSS can be divided into three steps, corresponding to the three trainable modules specified in Fig.~\ref{fig:overview}~(a). We first train the ResMLP of the Retrieval Engine using metric learning with the triplet loss defined in Eq.~(\ref{eq:triplet_loss}). During training, the positive pairs comprised samples that shared the same scanning stage but were drawn from different volumes to encourage generalization, whereas negative pairs were drawn either from different stages within the same volume or from other stages in different volumes. 

\par
Then, following the LLaVA paradigm~\cite{liu2023visual}~\cite{li2024llava}, the remaining modules are trained with two sequential steps: \textit{i). Pre-training the cross-modality projector for alignment:} Visual features extracted by the frozen PubMedCLIP encoder are aligned with textual descriptions through a two-layer MLP. \textit{ii). Instruction fine-tuning:} Multi-turn QA pairs from the VQA dataset are used to fine-tune the cross-modality MLP and the Vicuna backbone with LoRA adapters~\cite{hu2022lora}. Both stages are optimized with the same next-token prediction cross-entropy loss:
\begin{equation}
\mathcal{L}_{CE} = - \sum_{i=n+1}^{L} y_{l} \log P(x_{l} \mid z_{1:n}).
\end{equation}
\vspace{0.2cm}

\section{Experiments}
\label{sec:expres}
\par
To evaluate the performance of RAG-RUSS, we begin by evaluating the retrieval accuracy of the RAG component. Then, an ablation study is conducted to identify the individual contributions of the VLM and the RAG component, as well as to assess the impact of varying the number of retrieved examples.

\subsection{Implementation Details}~\label{sec:exp_imple}
\par
All experiments were conducted on a server with an NVIDIA A100 GPU (80 GB memory). Training was performed in 3 sequential stages using the AdamW optimizer. \textit{i) ResMLP module:} The ResMLP constructing the RAG component was trained for 100 epochs with batch sizes of 32 for training and 64 for validation, a learning rate of $3\times10^{-6}$, weight decay of $1\times10^{-5}$, and cosine decay scheduling with 10\% warm-up. \textit{ii) VLM pretraining:} The cross-modality projector in VLM was pretrained for 5 epochs on the training dataset without validation, using a batch size of 32, a learning rate of $1\times10^{-5}$, cosine decay scheduling with 3\% warm-up, and no weight decay. 
\textit{iii) ICL instruction tuning:} The MLP and LoRA adapters of the LLM were trained on the ICL dataset using the same settings as VLM pretraining.

\subsection{RAG Retrieval Accuracy}
\par
We evaluate the performance of RAG module using the Top@k metric,  defined as follows: for each query, Top@k is set to 1 if at least one relevant context example appears among the top-k retrieved results; otherwise, it is 0. The final score is reported as the average over all queries. Table~\ref{tab:acc_stage} summarizes the retrieval accuracy on the test set of dataset A. On average, the retriever attains 70.0\% Top@1 and 76.2\% Top@2, with 6.2\% gain from adding one extra context example. Stages with strong, distinctive visual cues perform best, such as \textit{Examine CCA proximal} (95.3\% Top@1) and \textit{Rotate to longitudinal view} (94.2\% Top@1), suggesting the retriever reliably identifies canonical appearances. In contrast, \textit{Transverse scan completed} is the most difficult stage, yields the lowest accuracy (37.3\% Top@2). While the ``return” phases remain challenging (\textit{Return to carotid bulb}: 59.7\% Top@2; \textit{Return completed}: 76.9\% Top@2), the largest Top@2 gains are observed for these stages (11.7\% and 9.9\%, respectively), indicating ambiguity that benefits from multiple references. In addition, \textit{Examine bifurcation} shows mid-level performance (66.7\% Top@1, 74.9\% Top@2), likely reflecting higher anatomical variability across cases. These results highlight two key observations: (i) using retrieval with k=2 improves robustness under uncertainty, and (ii) visually similar stages, especially during return and completion, require more discriminative training, such as harder negative mining, to enhance stage-level separation.

\begin{table}[!htbp]
\centering
\caption{Context Example Retrieval Accuracy}
\label{tab:rag_precision}
\resizebox{0.39\textwidth}{!}{%
\begin{tabular}{lcc}
\toprule
\textbf{Stage}
 & \textbf{Top@1 Acc. $\uparrow$} 
 & \textbf{Top@2 Acc.$\uparrow$} \\
\midrule
Examine CCA proximal       & 95.3\% & 96.2\% \\
Examine CCA distal         & 85.5\% & 89.3\% \\
Examine bifurcation        & 66.7\% & 74.9\% \\
Transverse scan completed  & 30.5\% & 37.3\% \\
Return to carotid bulb     & 48.0\% & 59.7\% \\
Return completed           & 67.0\% & 76.9\% \\
Rotate to longitudinal view & 94.2\% & 95.5\% \\
Longitudinal scan completed & 72.9\% & 80.0\% \\
\midrule
\textbf{Average} & 70.0\% & 76.2\% \\
\bottomrule
\end{tabular}
}
\vspace{-0.3cm}
\end{table}

\begin{table*}[!ht]
\centering
\caption{Stage-level recognition accuracy under different ablation settings.}
\label{tab:acc_stage}
\resizebox{0.95\textwidth}{!}{%
\begin{tabular}{lcccccccc}
\toprule
\textbf{Method}
 & \textbf{CCA proximal} 
 & \textbf{CCA distal } 
 & \textbf{Bifurcation } 
 & \textbf{Trans. completed } 
 & \textbf{Return completed } 
 & \textbf{Long. completed } 
 & \textbf{Avg acc. } \\ 
\midrule
\textbf{RAG only}            
 & 96.5\% & 82.3\% & \textbf{84.3\%} & 50.0\%  & 25.0\% & 50.0\% & 64.7\% \\
\textbf{VLM only}            
 & 92.6\% & 93.3\% & 69.2\% & 75.0\% & 50.0\% & 50.0\% & 71.7\% \\
\textbf{RAG-RUSS@1} 
 & 69.9\% & 92.1\% & 53.6\% & 50.0\% & \textbf{75.0\%} & \textbf{75.0\%} & 69.3\% \\
\textbf{RAG-RUSS@2} 
 & \textbf{97.8\%} & \textbf{95.0\%} & 81.7\% & \textbf{100.0\%} & 25.0\% & \textbf{75.0\%} & \textbf{79.1\%} \\
\bottomrule
\multicolumn{8}{l}{\scriptsize RAG-RUSS@1 and RAG-RUSS@2 denote configurations in which the RAG component retrieves one or two in-context learning examples, respectively, for the RAG-RUSS input.}\\
\multicolumn{8}{l}{\scriptsize Trans. completed: Transverse scan completed.~~~Long. completed: Longitudinal scan complete.~~~Avg. acc.: Average accuracy.}
\end{tabular}
}
\end{table*}

\begin{figure*}[!ht]
    \centering
    \includegraphics[width=.9\textwidth]{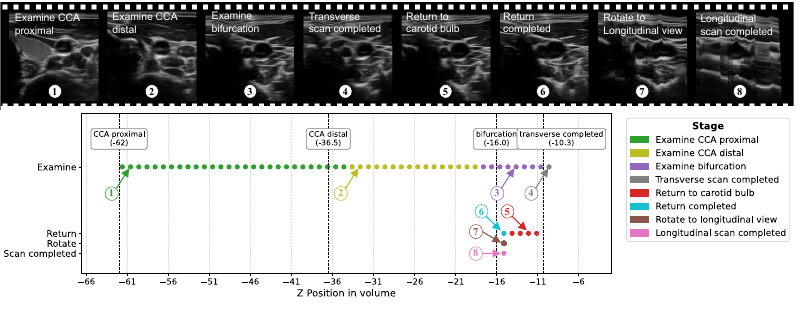}
    \vspace{-0.3cm}
    \caption{Visualization example of the closed-loop evaluation results of RAG-RUSS@2. The upper panel presents US images corresponding to eight stages predicted by the model. The lower panel illustrates the complete scanning trajectory together with the annotated scanning regions, providing a visualization of the model’s recognized progression through different stages.}
    \vspace{-0.4cm}
    \label{fig:result_visual}
\end{figure*}

\subsection{Ablation Study}
\par
To investigate the role of each component in RAG-RUSS and the impact of the number of retrieval examples, we conducted the following ablation experiment comparison: i). \textit{RAG Only:} Only RAG component was used to retrieve the most similar scanning context for stage prediction. ii). \textit{VLM Only:} This baseline relied solely on the VLM for prediction without retrieval augmentation. iii). \textit{RAG-RUSS@1} 
and iv). \textit{RAG-RUSS@2}: These configurations used the full framework, with the RAG component retrieving one or two in-context learning examples, respectively, as input to RAG-RUSS.



\par
Accordingly, we conducted a closed-loop evaluation to assess stage prediction accuracy during continuous scanning across four experimental settings on four unseen carotid artery volumes in the test set. The stage-wise accuracy results are presented in Table~\ref{tab:acc_stage}. For the stages \textit{Examine CCA proximal}, \textit{Examine CCA distal}, and \textit{Examine bifurcation}, accuracy is defined as the ratio between the duration of correctly recognized inferences and the annotated ground-truth duration. For \textit{Transverse scan completed}, accuracy is determined by whether the model successfully reaches or surpasses the labeled waypoint. For \textit{Return completed}, accuracy is defined as the predicted stage occurring within the annotated region between the bifurcations. Finally, for \textit{Longitudinal scan completed}, accuracy is assessed based on whether the longitudinal cross-section of the carotid artery is clearly visualized. An intuitive visualization for automatically completing a full-loop carotid examination on unseen human volumetric data using RAG-RUSS@2 is shown in Fig.~\ref{fig:result_visual}.

\par
The results are summarized in Table~\ref{tab:acc_stage}. Among all configurations, RAG Only performs the worst with an average accuracy of 64.7\%. In contrast, RAG-RUSS@2 (with two retrieved examples) achieves the highest average accuracy (79.1\%), demonstrating strong performance on anatomically well-defined stages such as \textit{CCA proximal} (97.8\%), \textit{CCA distal} (95.0\%), and \textit{Transverse scan completed} (100.0\%). These results suggest that combining VLM with multiple, diverse retrievals is particularly effective for recognizing clear anatomical landmarks and identifying stage completion events. The VLM Only configuration yields the most stable overall performance among the baselines, achieving an average accuracy of 71.7\%. It performs reliably on stages like \textit{CCA distal} (93.3\%) and \textit{Transverse scan completed} (75.0\%), but lacks the case-specific contextualization that retrieval provides. 

\subsection{Discussion}

\par
The experimental results demonstrate that RAG-RUSS@2 achieves the best overall performance, highlighting the effectiveness and benefits of augmenting VLMs with retrieval-based context for US scanning. However, a closer examination reveals that \textit{Return completed} emerges as the most challenging stage across all configurations. Interestingly, while RAG-RUSS@1 (with one retrieved example) achieves the highest accuracy on this stage (75.0\%), performance drops sharply to 25.0\% in RAG-RUSS@2, suggesting that retrieval-induced noise or conflicting contextual signals can hinder model performance. This highlights that more context is not always better, especially when temporal ambiguity is involved. Several limitations remain. RAG-RUSS requires significant computational resources, as longer inputs from multiple retrieved examples increase inference time. Additionally, the current system operates with discrete API-based control, rather than continuous action execution, which may constrain its adaptability in fine-grained scanning tasks. Furthermore, the approach has only been validated based on the prerecorded human data. Practical challenges in real scenarios, such as deformation, will pose additional challenges in future deployment. 
Nevertheless, RAG-RUSS demonstrates that integrating retrieval-augmented generation with vision–language reasoning can enable accurate, interpretable, and autonomous US scanning. This is an essential step toward enhancing trust and acceptance among both sonographers and patients.

\section{Conclusion}
\label{sec:conclusion}
\par
This work presents RAG-RUSS, an interpretable and autonomous framework for carotid US scanning that follows established clinical workflows and leverages VLM. 
To facilitate learning within a structured clinical workflow, we constructed a high-quality dataset from 32 human volumetric scans. Building on this dataset, we propose RAG-RUSS, which integrates an LLM, a vision foundation model, and a retrieval engine. In particular, the retrieval-augmented in-context learning strategy enhances adaptability and generalization while reducing the amount of training data required. Experimental results demonstrate that RAG-RUSS can autonomously perform full-stage carotid scanning in accordance with clinical protocols. Moreover, by providing explanations of the current image and the subsequent action, the system takes an important step toward developing trustworthy and acceptable RUSS for routine deployment.

%% file: main_icra.bbl
\begin{thebibliography}{10}
\providecommand{\url}[1]{#1}
\csname url@samestyle\endcsname
\providecommand{\newblock}{\relax}
\providecommand{\bibinfo}[2]{#2}
\providecommand{\BIBentrySTDinterwordspacing}{\spaceskip=0pt\relax}
\providecommand{\BIBentryALTinterwordstretchfactor}{4}
\providecommand{\BIBentryALTinterwordspacing}{\spaceskip=\fontdimen2\font plus
\BIBentryALTinterwordstretchfactor\fontdimen3\font minus \fontdimen4\font\relax}
\providecommand{\BIBforeignlanguage}[2]{{%
\expandafter\ifx\csname l@#1\endcsname\relax
\typeout{** WARNING: IEEEtran.bst: No hyphenation pattern has been}%
\typeout{** loaded for the language `#1'. Using the pattern for}%
\typeout{** the default language instead.}%
\else
\language=\csname l@#1\endcsname
\fi
#2}}
\providecommand{\BIBdecl}{\relax}
\BIBdecl

\bibitem{izadifar2017mechanical}
Z.~Izadifar, P.~Babyn, and D.~Chapman, ``Mechanical and biological effects of ultrasound: a review of present knowledge,'' \emph{Ultrasound in medicine \& biology}, vol.~43, no.~6, pp. 1085--1104, 2017.

\bibitem{sippel2011use}
S.~Sippel, K.~Muruganandan, A.~Levine, and S.~Shah, ``Use of ultrasound in the developing world,'' \emph{International journal of emergency medicine}, vol.~4, no.~1, p.~72, 2011.

\bibitem{jiang2023robotic}
Z.~Jiang, S.~E. Salcudean, and N.~Navab, ``Robotic ultrasound imaging: State-of-the-art and future perspectives,'' \emph{Medical image analysis}, vol.~89, p. 102878, 2023.

\bibitem{welleweerd2021out}
M.~K. Welleweerd, A.~G. de~Groot, V.~Groenhuis, F.~J. Siepel, and S.~Stramigioli, ``Out-of-plane corrections for autonomous robotic breast ultrasound acquisitions,'' in \emph{2021 IEEE International Conference on Robotics and Automation (ICRA)}.\hskip 1em plus 0.5em minus 0.4em\relax IEEE, 2021, pp. 12\,515--12\,521.

\bibitem{ma2021autonomous}
X.~Ma, Z.~Zhang, and H.~K. Zhang, ``Autonomous scanning target localization for robotic lung ultrasound imaging,'' in \emph{2021 IEEE/RSJ International Conference on Intelligent Robots and Systems (IROS)}.\hskip 1em plus 0.5em minus 0.4em\relax IEEE, 2021, pp. 9467--9474.

\bibitem{huang2023motion}
D.~Huang, Y.~Bi, N.~Navab, and Z.~Jiang, ``Motion magnification in robotic sonography: enabling pulsation-aware artery segmentation,'' in \emph{2023 IEEE/RSJ International Conference on Intelligent Robots and Systems (IROS)}.\hskip 1em plus 0.5em minus 0.4em\relax IEEE, 2023, pp. 6565--6570.

\bibitem{huang2024robotassist}
D.~Huang, C.~Yang, M.~Zhou, A.~Karlas, N.~Navab, and Z.~Jiang, ``Robot-assisted deep venous thrombosis ultrasound examination using virtual fixture,'' \emph{IEEE Transactions on Automation Science and Engineering}, vol.~22, pp. 381--392, 2024.

\bibitem{wang2024autonomous}
Z.~Wang, Y.~Han, B.~Zhao, H.~Xie, L.~Yao, B.~Li, M.~Q.-H. Meng, and Y.~Hu, ``Autonomous robotic system for carotid artery ultrasound scanning with visual servo navigation,'' \emph{IEEE Transactions on Medical Robotics and Bionics}, 2024.

\bibitem{huang2024robot}
Q.~Huang, B.~Gao, and M.~Wang, ``Robot-assisted autonomous ultrasound imaging for carotid artery,'' \emph{IEEE Transactions on Instrumentation and Measurement}, vol.~73, pp. 1--9, 2024.

\bibitem{bi2024machine}
Y.~Bi, Z.~Jiang, F.~Duelmer, D.~Huang, and N.~Navab, ``Machine learning in robotic ultrasound imaging: Challenges and perspectives,'' \emph{Annu. Rev. Control. Robotics Auton.}, vol.~7, 2024.

\bibitem{huang2021towards}
Y.~Huang, W.~Xiao, C.~Wang, H.~Liu, R.~Huang, and Z.~Sun, ``Towards fully autonomous ultrasound scanning robot with imitation learning based on clinical protocols,'' \emph{IEEE Robotics and Automation Letters}, vol.~6, no.~2, pp. 3671--3678, 2021.

\bibitem{jiang2024intelligent}
Z.~Jiang, Y.~Bi, M.~Zhou, Y.~Hu, M.~Burke, and N.~Navab, ``Intelligent robotic sonographer: Mutual information-based disentangled reward learning from few demonstrations,'' \emph{The International Journal of Robotics Research}, vol.~43, no.~7, pp. 981--1002, 2024.

\bibitem{su2024fully}
K.~Su, J.~Liu, X.~Ren, Y.~Huo, G.~Du, W.~Zhao, X.~Wang, B.~Liang, D.~Li, and P.~X. Liu, ``A fully autonomous robotic ultrasound system for thyroid scanning,'' \emph{Nat. Comm.}, vol.~15, no.~1, p. 4004, 2024.

\bibitem{li2021autonomous}
K.~Li, J.~Wang, Y.~Xu, H.~Qin, D.~Liu, L.~Liu, and M.~Q.-H. Meng, ``Autonomous navigation of an ultrasound probe towards standard scan planes with deep reinforcement learning,'' in \emph{ICRA}.\hskip 1em plus 0.5em minus 0.4em\relax IEEE, 2021, pp. 8302--8308.

\bibitem{song2025intelligent}
T.~Song, F.~Li, Y.~Bi, A.~Karlas, A.~Yousefi, D.~Branzan, Z.~Jiang, U.~Eck, and N.~Navab, ``Intelligent virtual sonographer (ivs): Enhancing physician-robot-patient communication,'' \emph{arXiv preprint arXiv:2507.13052}, 2025.

\bibitem{kim2025srt}
J.~W. Kim, J.-T. Chen, P.~Hansen, L.~X. Shi, A.~Goldenberg, S.~Schmidgall, P.~M. Scheikl, A.~Deguet, B.~M. White, D.~R. Tsai \emph{et~al.}, ``Srt-h: A hierarchical framework for autonomous surgery via language-conditioned imitation learning,'' \emph{Science robotics}, vol.~10, no. 104, p. eadt5254, 2025.

\bibitem{roy2021machine}
N.~Roy, I.~Posner, T.~Barfoot, P.~Beaudoin, Y.~Bengio, J.~Bohg, O.~Brock, I.~Depatie, D.~Fox, D.~Koditschek \emph{et~al.}, ``From machine learning to robotics: Challenges and opportunities for embodied intelligence,'' \emph{arXiv preprint arXiv:2110.15245}, 2021.

\bibitem{liu2025aligning}
Y.~Liu, W.~Chen, Y.~Bai, X.~Liang, G.~Li, W.~Gao, and L.~Lin, ``Aligning cyber space with physical world: A comprehensive survey on embodied ai,'' \emph{IEEE/ASME Transactions on Mechatronics}, 2025.

\bibitem{driess2023palm}
D.~Driess, F.~Xia, M.~S. Sajjadi, C.~Lynch, A.~Chowdhery, A.~Wahid, J.~Tompson, Q.~Vuong, T.~Yu, W.~Huang \emph{et~al.}, ``Palm-e: An embodied multimodal language model,'' 2023.

\bibitem{ng2025endovla}
C.~K. Ng, L.~Bai, G.~Wang, Y.~Wang, H.~Gao, K.~Yuan, C.~Jin, T.~Zeng, and H.~Ren, ``Endovla: Dual-phase vision-language-action model for autonomous tracking in endoscopy,'' \emph{arXiv preprint arXiv:2505.15206}, 2025.

\bibitem{chen2025uspilot}
M.~Chen, S.~Fan, G.~Cao, Y.-h. Liu, and H.~Liu, ``Uspilot: An embodied robotic assistant ultrasound system with large language model enhanced graph planner,'' \emph{arXiv preprint arXiv:2502.12498}, 2025.

\bibitem{jiang2025towards}
H.~Jiang, A.~Zhao, Q.~Yang, X.~Yan, T.~Wang, Y.~Wang, N.~Jia, J.~Wang, G.~Wu, Y.~Yue \emph{et~al.}, ``Towards expert-level autonomous carotid ultrasonography with large-scale learning-based robotic system,'' \emph{Nature Communications}, vol.~16, no.~1, p. 7893, 2025.

\bibitem{yuan2024rag}
J.~Yuan, S.~Sun, D.~Omeiza, B.~Zhao, P.~Newman, L.~Kunze, and M.~Gadd, ``Rag-driver: Generalisable driving explanations with retrieval-augmented in-context multi-modal large language model learning,'' in \emph{Robotics: Science and Systems}, 2024.

\bibitem{lasso2014plus}
A.~Lasso, T.~Heffter, A.~Rankin, C.~Pinter, T.~Ungi, and G.~Fichtinger, ``Plus: open-source toolkit for ultrasound-guided intervention systems,'' \emph{IEEE transactions on biomedical engineering}, vol.~61, no.~10, pp. 2527--2537, 2014.

\bibitem{hu2022lora}
E.~J. Hu, Y.~Shen, P.~Wallis, Z.~Allen-Zhu, Y.~Li, S.~Wang, L.~Wang, W.~Chen \emph{et~al.}, ``Lora: Low-rank adaptation of large language models.'' \emph{ICLR}, vol.~1, no.~2, p.~3, 2022.

\bibitem{zheng2023judging}
L.~Zheng, W.-L. Chiang, Y.~Sheng, S.~Zhuang, Z.~Wu, Y.~Zhuang, Z.~Lin, Z.~Li, D.~Li, E.~Xing \emph{et~al.}, ``Judging llm-as-a-judge with mt-bench and chatbot arena,'' \emph{Advances in neural information processing systems}, vol.~36, pp. 46\,595--46\,623, 2023.

\bibitem{eslami2021does}
S.~Eslami, G.~De~Melo, and C.~Meinel, ``Does clip benefit visual question answering in the medical domain as much as it does in the general domain?'' \emph{arXiv preprint arXiv:2112.13906}, 2021.

\bibitem{ruckert2024rocov2}
J.~R{\"u}ckert, L.~Bloch, R.~Br{\"u}ngel, A.~Idrissi-Yaghir, H.~Sch{\"a}fer, C.~S. Schmidt, S.~Koitka, O.~Pelka, A.~B. Abacha, A.~G.~Seco~de Herrera \emph{et~al.}, ``Rocov2: Radiology objects in context version 2, an updated multimodal image dataset,'' \emph{Scientific Data}, vol.~11, no.~1, p. 688, 2024.

\bibitem{liu2023visual}
H.~Liu, C.~Li, Q.~Wu, and Y.~J. Lee, ``Visual instruction tuning,'' \emph{Advances in neural information processing systems}, vol.~36, pp. 34\,892--34\,916, 2023.

\bibitem{li2024llava}
F.~Li, R.~Zhang, H.~Zhang, Y.~Zhang, B.~Li, W.~Li, Z.~Ma, and C.~Li, ``Llava-next-interleave: Tackling multi-image, video, and 3d in large multimodal models,'' \emph{arXiv preprint arXiv:2407.07895}, 2024.

\end{thebibliography}
